\pdfoutput=1
\documentclass[10pt,twocolumn]{article}

\usepackage[a4paper,top=1.55cm,bottom=1.7cm,left=1.65cm,right=1.65cm]{geometry}
\usepackage[T1]{fontenc}
\usepackage[utf8]{inputenc}
\usepackage{newtxtext}
\usepackage{microtype}
\usepackage{inconsolata}
\usepackage{graphicx}
\usepackage{booktabs}
\usepackage{multirow}
\usepackage{amsmath}
\usepackage{amssymb}
\usepackage{float}
\usepackage{fvextra}
\usepackage{array}
\usepackage{natbib}
\usepackage{titlesec}
\usepackage{fancyhdr}
\usepackage{caption}
\usepackage{tcolorbox}
\usepackage{fontawesome5}
\usepackage{enumitem}
\usepackage{hyperref}

\definecolor{accent}{HTML}{075E5A}
\definecolor{accentlink}{HTML}{0B9FAD}
\definecolor{abstractbg}{HTML}{EEF7F6}
\definecolor{rulegray}{HTML}{333333}
\definecolor{muted}{HTML}{555555}

\hypersetup{
  colorlinks=true,
  linkcolor=accent,
  citecolor=accentlink,
  urlcolor=accentlink,
  pdfauthor={Xiao Fan, Jingyuan Li, Hongbin Guo, Yubo Han, Yi Zhang},
  pdftitle={Provenance Before Prose: Claim-Locked Reporting for Statistical Text Generation}
}
\setlist[itemize]{leftmargin=1.3em,itemsep=2pt,topsep=3pt}

\titleformat{\section}
  {\large\bfseries\color{accent}}
  {\thesection.}{0.55em}{}
\titleformat{name=\section,numberless}
  {\large\bfseries\color{accent}}
  {}{0pt}{}
\titleformat{\subsection}
  {\normalsize\bfseries\color{accent}}
  {\thesubsection}{0.55em}{}
\titlespacing*{\section}{0pt}{11pt}{5pt}
\titlespacing*{\subsection}{0pt}{8pt}{3pt}

\fancypagestyle{firstpage}{%
  \fancyhf{}
  \fancyfoot[C]{\thepage}
  
}

\newcommand{\preprinttopmatter}{%
\begin{minipage}{\textwidth}
  \thispagestyle{firstpage}
  \noindent
  {\small\bfseries Xidian University}\hfill{\small\itshape Accepted to EMNLP: August 2026}\\[-0.35em]
  \color{rulegray}\rule{\textwidth}{0.7pt}\color{black}

  \vspace{0.75em}
  {\fontsize{21.5}{24}\selectfont\bfseries\color{accent}
  Provenance Before Prose:\\[-0.05em]
  Claim-Locked Reporting for Statistical Text Generation\par}

  \vspace{0.7em}
  {\large\bfseries
  Xiao Fan,\quad Jingyuan Li\textsuperscript{*},\quad Hongbin Guo,\quad
  Yubo Han,\quad Yi Zhang\textsuperscript{*}\par}
  \vspace{0.22em}
  {\normalsize Xidian University\par}
  {\small\texttt{\{xiufan, hongbin, hanyubo\}@stu.xidian.edu.cn}\quad
  \texttt{\{lijingyuan, yizhang\}@xidian.edu.cn}\par}
  \vspace{0.28em}
  {\small\textsuperscript{*}Corresponding authors.\par}

  \vspace{0.48em}
  {\small
  \href{https://github.com/XiuFan719/Claim-Locked-Reporting}{\faGithub\ \textbf{Code}}\par}
  \vspace{0.25em}
  \color{rulegray}\rule{\textwidth}{0.7pt}\color{black}

  \vspace{0.72em}
  \begin{tcolorbox}[
    colback=abstractbg,
    colframe=abstractbg,
    boxrule=0pt,
    arc=2.2mm,
    left=3.5mm,right=3.5mm,top=2.8mm,bottom=2.8mm
  ]
  \textbf{Abstract}\quad Large language models (LLMs) can fluently verbalize statistical evidence, yet statistical reports can still drift numerical values, invert effect directions, or restate thresholded contrasts as categorical effects. We frame these failures as a control problem: the evidence-bearing content of a scientific report should be fixed by structured statistical results rather than sampled during prose generation. We therefore use cross-run reproducibility to stress-test whether report-visible numbers and claims are bound before prose generation. Existing controls operate at the text or slot level; a deterministic hybrid template reproduces only $61.1\%$ of report-visible numerical content across seeds because the LLM still selects which findings and numbers the template renders. We propose \emph{claim-locked reporting}, a provenance-before-prose protocol that fixes the evidence source, numbers, direction, and allowed language strength of each reportable claim before the LLM writes connective wording. Across fMRI functional-connectivity reporting and randomized controlled trial reporting on Evidence Inference 2.0, claim-locked reporting improves reproducibility over the hybrid template by $37.4$ and $20.5$ points, respectively. Blinded human audits support the observed direction-preservation and governance trends. In an fMRI cost analysis with DeepSeek, claim-locked reporting also yields the lowest observed token use and median generation latency. Code is available at \url{https://github.com/XiuFan719/Claim-Locked-Reporting}.
  \end{tcolorbox}
  \vspace{0.55em}
\end{minipage}%
}

\begin{document}
\twocolumn[\preprinttopmatter]

\section{Introduction}
Consider a functional magnetic resonance imaging (fMRI) report produced by a large language model (LLM). Before any prose is written, the statistical results are already fixed by the analysis. Yet across runs, the same results can still lead to different reports. One report may include a finding that another omits, report a different numerical detail, or describe the same group effect with stronger or weaker wording.
Such variation changes the evidence-bearing content of the report. When the underlying results have not changed, the set of reported findings and numbers, together with their interpretive strength, should remain stable across runs.

LLMs are increasingly used to draft scientific and clinical reports, and much of the current reliability framing asks whether generated text is faithful to an observable or retrievable source, such as an image, a patient record, or a retrieved passage~\citep{ref_vanveen,ref_maira2,ref_medpalmm,ref_singhal}. For LLM-generated statistical reports, the reliability question is different. Functional connectivity (FC) analyses, randomized controlled trial (RCT) summaries, and epidemiological reports are written from statistical results that have already been computed: counts, effect estimates, directions, thresholds, covariate conditions, or evidence spans~\citep{ref_bullmore,ref_zalesky,ref_marek}. A report may use numbers or terms that appear in the source while still changing the statistical claim it conveys, for example by drifting a number, reversing a direction, or overstating the strength of an association. We refer to this family of errors as \emph{statistical claim distortion}. Reliability in statistical reporting requires preserving not only source-supported content, but also the statistical meaning of that content.

Existing controls intervene at different points in the generation pipeline. For clarity, we group them into two levels. Text-level control includes prompting, retrieval~\citep{ref_rag}, structured output~\citep{ref_outlines,ref_guidance}, and post-hoc verification~\citep{ref_self_check}. These methods can constrain the generation process or output form, but the model still decides which statistical inferences to state and how strongly to state them. Slot-level control combines selected fields with deterministic surface realization~\citep{ref_reiter_dale,ref_gatt_krahmer}. It ensures that a selected field is rendered consistently, but it does not decide which field should be selected in the first place. When the LLM chooses the slots, the reported numbers, directions, and language strength can still vary across runs.

This leaves a claim-level gap. Existing controls can constrain the text or render selected slots, but they do not fix the reportable statistical claim before rendering. We address this gap with claim-locked reporting, a third control level that binds each reportable claim to its evidence source, numbers, effect direction, and allowed language strength before the LLM writes. A claim builder converts the structured evidence record into a claim ledger; a risk auditor and policy controller determine which claims are admissible and how strongly they may be stated; a deterministic renderer emits numbers, tables, entity labels, and direction tags from the ledger; and the LLM writes only connective prose around the locked claims. The protocol changes what the model is allowed to decide.
We evaluate claim-locked reporting across FC reporting and RCT reporting using complementary automatic and human analyses of report stability, statistical correctness, and claim governance. Additional experiments examine component contributions and practical generation cost.
Our contributions are summarized as follows:

\begin{itemize}
	\item We frame statistical claim distortion as a reliability problem in LLM-generated statistical reporting, and use cross-run reproducibility as a stress test of whether evidence-bearing content is fixed before prose generation.
	
	\item We propose claim-locked reporting, a provenance-before-prose protocol that binds each reportable claim to its evidence source, numbers, direction, and allowed language strength before prose generation.
	
	\item We evaluate the protocol on obesity-related FC reporting and RCT reporting on Evidence Inference 2.0 against grounded baselines, supplemented by component analysis and blinded human audits. Claim-locked reporting improves reproducibility over the hybrid template by $37.4$ and $20.5$ points in the two settings, respectively.
\end{itemize}

\section{Related Work}

\paragraph{Statistical reporting versus faithful generation.} Classical natural language generation separates content planning from surface realization~\citep{ref_reiter_dale,ref_gatt_krahmer}: the system first decides what to communicate and then how to express it. Neural data-to-text generation later learned content selection and realization jointly from structured records~\citep{ref_lebret,ref_wiseman}, and explicitly modeling content selection and planning was subsequently shown to improve generation quality~\citep{ref_puduppully}. Statistical reporting, however, exposes a distinction this line of work does not make: selecting the correct records does not guarantee a correct statistical claim. The same supported evidence can still be verbalized with the wrong direction, stripped of an adjustment condition, or stated with unjustified strength. LLM-era faithfulness evaluation operates on the generated output. FActScore~\citep{ref_factscore} decomposes text into atomic facts and evaluates whether they are supported, while SelfCheckGPT~\citep{ref_self_check} uses consistency across sampled generations as a signal of hallucination; broader surveys similarly organize failures around factual support and consistency~\citep{ref_hallucination_survey,ref_huang_hallucination}. These methods evaluate properties of generated content rather than explicitly representing the statistical claim that the content supports. A report may therefore contain individually supported facts while changing their joint statistical meaning. Such errors concern what the report claims, not merely whether its individual facts appear in the source. We therefore treat these failures as a control problem rather than relying solely on post-generation detection.
\paragraph{Constrained generation for statistical reporting.}
Existing approaches improve grounding or constrain generation through retrieval-augmented generation (RAG)~\citep{ref_rag},
constrained decoding and structured generation~\citep{ref_outlines,ref_guidance}, and post-hoc verification~\citep{ref_self_check}.
These approaches constrain evidence access, output structure, or generated text. Statistical reporting, however, requires the numerical value, direction, inferential scope, and interpretive strength of a claim to remain jointly tied to its supporting evidence. Claim-locked reporting therefore treats the provenance-bound statistical claim, rather than text or selected fields, as the
unit of control.

\section{Claim-Locked Reporting}
\label{sec:method}

\subsection{Statistical Evidence Record}

Claim-locked reporting starts from a structured evidence record rather than from raw imaging or trial documents. The record is the fixed statistical input to the claim builder and contains only results that have already been computed by a domain analysis pipeline.

In the FC instantiation, subject-level connectivity matrices and phenotypic variables are converted into this record by a standard edge-wise statistical analysis. We fit an edge-wise generalized linear model (GLM) contrasting binary group labels against demographic covariates, followed by Benjamini--Hochberg false-discovery-rate (FDR) control across $M=N(N-1)/2$ region-of-interest (ROI) pairs~\citep{ref_bh}. A parallel model additionally adjusts for body mass index (BMI), the continuous covariate used to derive the group labels, providing the covariate-absorption stress test (\S\ref{sec:bmi}). The resulting record stores FDR-surviving edge counts, lobe-pair summaries, hub descriptions, representative edges, optional behavior associations, and covariate-adjusted sensitivity results. 
For FC literature-context claims, an LLM-assisted literature-support label initializes the wording-strength ceiling and is frozen before report generation.

In the RCT instantiation, the evidence record is provided by Evidence Inference 2.0~\citep{ref_ei2}. Each instance contains an intervention, comparator, outcome, direction label, and an evidence span containing the supporting numerical content.

\subsection{Locked Statistical Claim}
\label{sec:components}

Three components construct and constrain the locked claims before report generation: the claim builder, risk auditor, and policy controller. They operate over a shared data structure, the claim ledger.

\paragraph{Claim builder.}
The claim builder converts the evidence record into the claim ledger, a set of typed claims. Each claim stores a claim type, canonical text, one or more evidence pointers into the record, the numerical fields it reports, an effect direction when applicable, a literature-support level, and an allowed language strength. A claim is instantiated only when its required fields resolve to evidence pointers. As a result, a statement such as ``the analysis demonstrates a causal mechanism'' is never instantiated unless such a claim is explicitly supported by the evidence record. Because downstream rendering and writing are driven by the ledger, this step fixes the set of inferences the report may contain before the LLM is involved.

\paragraph{Risk auditor.}
The risk auditor is a read-only layer. It inspects each claim and emits a per-claim audit record without modifying the ledger. Each claim is checked against a pre-specified set of statistical-reporting risk categories. The taxonomy is organized around five evidence-bearing properties that must be preserved when a statistical claim is verbalized: numerical support, direction preservation, entity/source support, inferential scope, and interpretive strength. Domain-specific audit rules instantiate these properties using risks documented in the corresponding scientific literature. In FC, these include motion-related confounding~\citep{ref_power}, post-selection and circular-analysis effects~\citep{ref_vul,ref_kriegeskorte}, and replicability concerns~\citep{ref_marek,ref_narps}.

The categories are domain-specific, but they instantiate the same auditor interface: a fixed trigger over structured evidence fields maps to a policy action before prose generation. In FC, the triggers operate over edges, hubs, ROI names, covariate-adjusted models, and brain--behavior associations. In RCT reporting, they operate over Evidence Inference fields, including intervention, comparator, outcome, direction label, and evidence span. Appendix~\ref{app:risk-tax} summarizes how the shared control targets are instantiated in the two domains.

Direction inversion is handled by binding each directional claim to a ledger-stored direction field rather than to a free-text label, so report-visible direction tags are rendered directly from that field. The auditor also rejects any claim whose evidence pointer fails to resolve against the stage outputs, enforcing the provenance requirement before writing. For each flagged claim, it records the required action to be applied by the policy controller.
\begin{figure}[t]
	\centering
	\includegraphics[width=\linewidth]{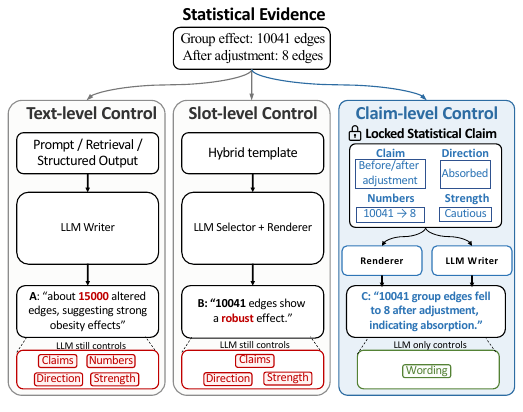}
	\caption{Three levels of control for statistical text generation. Text-level methods, including prompting, retrieval, and structured output, leave claims, numbers, directions, and language strength to the LLM. The hybrid template adds slot-level rendering, but the LLM still selects which claims and numbers fill the slots. Claim-locked reporting fixes the evidence source, numbers, direction, and allowed language strength before prose generation; the LLM controls only connective wording.}
	\label{fig:control}
\end{figure}

\paragraph{Policy controller.}
The policy controller turns audit records into rendering and writing constraints through two operations. First, it applies a monotone strength downgrade: a flagged claim's allowed language strength can only be lowered, never raised, and a forbidden claim is marked non-renderable and excluded from every report block. Second, it accumulates writer constraints, namely per-category natural-language directives that are later passed to the LLM as explicit instructions. For group-covariate circularity, for example, the constraint forbids describing the binary group label as an effect independent of the continuous covariate. The auditor decides which claims carry a risk; the policy controller decides what may be said about a flagged claim and whether it remains renderable. The same policy interface applies in the RCT setting, where risk tags are triggered over Evidence Inference fields rather than FC-specific fields.

\subsection{Deterministic Renderer and LLM Writer}

Report writing has two channels. The deterministic renderer emits all numerical content (GLM counts, hub tables, representative edges, limitations) directly from evidence pointers. The LLM writes only short connective paragraphs under the active policy constraints; if its output envelope is malformed or omits required sections, generation fails rather than producing a partial report. An audit outcome never asks the LLM to recompute statistics; it only restricts how a pre-computed claim may be verbalized.

Because both systems use deterministic rendering, their key distinction lies in the renderer input. In the hybrid template, the LLM reads the evidence record, selects values and list entries for predefined slots, and the renderer verbalizes those LLM-selected slots. In claim-locked reporting, the renderer receives ledger-bound claims whose source, numbers, direction, and allowed strength were already fixed by the builder and policy controller. The LLM still contributes local coherence, but it no longer decides which statistical content becomes reportable, nor how strongly it is stated.

\subsection{Worked Example: A Locked Claim}
\label{sec:worked}

We illustrate the pipeline with the group-effect claim from the in-house FC cohort. The evidence record contains two GLM stages: the primary model reports $10{,}041$ FDR-surviving group-effect edges, whereas the BMI-adjusted model reports $8$, a $99.9\%$ collapse. The claim builder therefore creates a single group-effect summary claim pointing to both stages, with the fields $10{,}041$, $8$, and $99.9\%$, a direction field indicating absorption after adjustment, and an initial language-strength limit from the literature-support label. The risk auditor attaches the group-covariate circularity tag, and the policy controller downgrades the claim to cautious language and forbids phrases such as ``independent obesity effect'' and ``obesity-specific mechanism''. The renderer emits the before-and-after counts directly from the ledger, while the LLM writes only cautious connective prose stating that the categorical group label largely reflects continuous BMI variation. Thus, unlike a free-form writer or a hybrid template rendering an LLM-selected primary count, claim-locked reporting fixes the claim, direction, and allowed strength before writing. Appendix~\ref{app:ledger} gives the corresponding ledger entry.

\subsection{Evaluation Axes}
\label{sec:audit}

We define three evaluation axes to test whether a statistical-report generator preserves fixed evidence as it moves from a structured record to prose. The evaluation definitions are fixed before evaluation. Implementation details for the automatic metrics are given in Appendix~\ref{app:audit-rules}, and the human audits are described in Appendix~\ref{app:human-audits}. We report the three axes separately because they characterize distinct failure modes: cross-run stability of report-visible content, governance of rhetorical strength, and per-run correctness of numbers and directions.

\paragraph{Evaluation as protocol testing.}
The evaluation focuses on preservation of fixed statistical evidence through the reporting pipeline. In open-ended factuality evaluation, the model may choose which facts to mention, and the output is later checked against a reference. In claim-locked reporting, the intended behavior is different: claim selection, numerical realization, direction framing, and strength calibration should no longer be unconstrained LLM choices. The axes below therefore test whether each control level keeps these properties fixed at the report surface.

\begin{figure}[t]
	\centering
	\includegraphics[width=\linewidth]{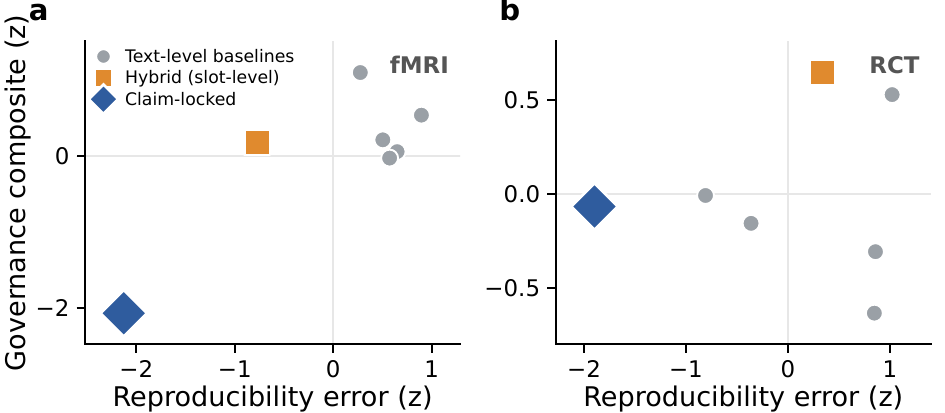}
	\caption{Summary landscape of the reproducibility-error and governance composites per method; lower is better on both axes, which are reported separately rather than merged into a single score. (a) fMRI; (b) RCT.}
	\label{fig:pareto}
\end{figure}

\paragraph{Cross-run reproducibility.}
Cross-run reproducibility measures whether the same evidence yields the same report-visible numerical content under different seeds and providers. We compute pooled cross-seed Jaccard overlap ($J$) of report-visible numerical tokens, reported as a percentage. Numerical matching uses absolute tolerance $10^{-3}$ or relative tolerance $5\%$, with a heuristic filter for section, table, and ordinal numbers. This axis is used as a stress test for whether numerical content is fixed before prose generation or resampled during generation. For visualization only, Figure~\ref{fig:pareto} plots $z(1-J)$, z-scored across methods.

\paragraph{Claim governance.} 
Claim governance evaluates whether risk-flagged claims are expressed with appropriate rhetorical strength in the final report. We report two direct measures. \textit{Strong} counts sentences that contain strong assertion terms without a hedge in the same sentence, with lower values indicating fewer over-strong statements. \textit{Hedge} measures the use of cautionary language for claims flagged by the risk taxonomy. For the summary visualization in Figure~\ref{fig:pareto}, we form the governance composite as the unweighted mean of $z(\mathrm{Strong})$ and $-z(\mathrm{Hedge})$, with lower values indicating stronger compliance.

\paragraph{Per-run correctness.}
The numerical audit flag (Num) marks report-visible numbers without an evidence match. Direction correctness is evaluated separately by setting. For RCT, a blinded human audit measures whether the generated report preserves, inverts, or leaves unresolved the gold direction (\S\ref{sec:human}). For FC, a rule-based lexical check compares direction wording against the ledger-stored direction field and is used only as a supplementary diagnostic because lexical direction checks are brittle under negation and comparator phrasing.

The key controlled contrast is between the hybrid template and claim-locked reporting. Both systems render deterministically, so the comparison tests whether reliability improves when the control unit moves from LLM-selected slots to evidence-bound claims. Scalar faithfulness metrics are treated as diagnostic, while blinded human audits are used to check whether the main governance and direction trends are visible to independent annotators.
\section{Experimental Setup}
\label{sec:setup}
We evaluate on two statistical-reporting settings that stress different parts of the same control problem: cohort-level neuroimaging reports and clinical-trial summaries.

\paragraph{fMRI FC setting.} We use obesity-related FC reporting as a representative cohort-level statistical-reporting task: the same phenotypic variables can serve as scientific descriptors, continuous covariates, and confounds, so a report must preserve not only numbers and directions but the inferential scope of each claim after covariate adjustment. We evaluate two cohorts: an obesity-focused institutional cohort of $428$ participants ($248$ obese, $180$ normal-weight) and the public Human Connectome Project (HCP) S1200 cohort~\citep{ref_hcp} with $712$ adults ($232$ obese, $480$ normal-weight) after excluding the overweight stratum. Both include demographic and clinical phenotypes as regression covariates, use $g{=}0$ for BMI $<25$ and $g{=}1$ for BMI $\geq 30$ following World Health Organization (WHO) thresholds~\citep{ref_who}, and use the Brainnetome-246 parcellation~\citep{ref_bna}. The evaluation spans seven methods across two cohorts, two writer providers, and five seeds, yielding $20$ cells per method.

\paragraph{RCT setting.} We use Evidence Inference 2.0~\citep{ref_ei2}, a public BioNLP benchmark of clinical-trial questions paired with full-text articles and human-annotated evidence spans. We sample $200$ records deduplicated by source article, retaining only records with annotator consensus, a non-null direction label, and at least two numerical tokens in the evidence span. This yields $112$ significantly increased and $88$ significantly decreased records; the no-significant-difference class is excluded because a direction-inversion audit is undefined when the gold direction is null. The RCT results are therefore a directional-claim benchmark. Each method runs with two seeds and two providers, yielding $800$ cells per method.

\paragraph{Human and scalar sub-samples.} For the scalar-metric comparison (\S\ref{sec:scalar}), we evaluate FActScore and SelfCheckGPT on a $30$-record RCT sub-sample; cross-metric comparison is restricted to RCT because FActScore's evidence-span ground truth has no direct fMRI analogue. The same $30$ records are used for the human direction audit, yielding $30\times7\times2\times2=840$ generation cells. FActScore requires one LLM judge call per generation; SelfCheckGPT follows the original protocol with $N=10$. The fMRI governance sanity check uses matched report snippets from free-form, hybrid template, and claim-locked reporting; four reviewers count two failure types, unhedged strong-language and unsupported-content violations, without seeing method identity.

\paragraph{Methods and fairness.} Five grounded baselines combine evidence injection with one structural or post-hoc control each: Free-form, Prompt-only (faithfulness instruction), Structured (output schema), Retrieval (evidence-block citation grounding), and Post-hoc verifier (generate then check). The hybrid template implements slot-level control through deterministic rendering but still lets the LLM select which fields to render, isolating slot-level control from claim-level control. Claim-locked reporting implements the full protocol. We use two writer providers, Moonshot \texttt{kimi-k2.6} (temperature $1.0$) and DeepSeek \texttt{deepseek-v4-pro} (temperature $0.7$). All methods receive the identical structured evidence record for a given cohort, provider, and seed; they differ only in the control point applied between record and report. Appendix~\ref{app:baselines} gives the per-method control envelope.

\section{Results}

\subsection{Main Controlled Contrast: Slot Rendering Is Not Enough}
\label{sec:main}

\begin{table}[t]
\caption{Main fMRI comparison ($n=20$ cells per row). Hedge is risk-conditioned; Num is a numerical audit flag. Bold = best among the seven methods (Repro., Strong, Num.).}
\label{tab:main}
\centering
\footnotesize
\setlength{\tabcolsep}{4pt}
\begin{tabular}{lrrrr}
\toprule
Method & Repro.\,$\uparrow$ & Hedge & Strong\,$\downarrow$ & Num.\,$\downarrow$ \\
\midrule
Free-form          & 32.3 & 0.20 & 2.95 & 0.55 \\
Prompt-only        & 26.0 & 0.22 & 1.95 & 0.55 \\
Structured         & 22.0 & 0.23 & 1.85 & 0.35 \\
Retrieval          & 24.1 & 0.21 & 1.50 & 0.20 \\
Post-hoc verifier  & 15.2 & 0.25 & 2.75 & 0.90 \\
Hybrid template    & 61.1 & 0.25 & 2.25 & \textbf{0.00} \\
\textbf{Claim-locked} & \textbf{98.5} & 0.39 & \textbf{0.75} & \textbf{0.00} \\
\bottomrule
\end{tabular}
\end{table}

Table~\ref{tab:main} and Figure~\ref{fig:pareto}(a) compare the three control levels of Figure~\ref{fig:control}: text-level control remains unstable, slot-level control improves stability, and claim-locked reporting makes evidence-bearing content nearly invariant across seeds.

\paragraph{Cross-run reproducibility.} Grounded baselines reach $15.2$--$32.3\%$ cross-seed reproducibility, indicating that report-visible numerical content remains a sampling outcome under these controls. Deterministic rendering is a major lever but not a sufficient one: the hybrid template reaches only $61.1\%$, because the LLM still selects which content enters the template. Claim-locked reporting reaches $98.5\%$. A ledger-rendered system is expected to reach a high absolute value, so the informative quantity is the $37.4$-point gap over the hybrid template, which also renders deterministically; the gap isolates the effect of additionally fixing which evidence-bound claims are renderable. A paired bootstrap over matched units (20{,}000 resamples) gives a 95\% confidence interval (CI) of $[+15.1,+59.7]$ points for this difference. The pattern also appears across both writer providers: the pooled metric reaches $98.5\%$ for fMRI and $100.0\%$ for RCT although the two providers have different numerical and hedging priors. The same control-level pattern holds on the public HCP cohort alone: claim-locked reaches $98.0\%$ reproducibility versus $62.0\%$ for the hybrid template and $46.0\%$ for free-form generation.

\paragraph{Governance and numerical flags.} The remaining columns of Table~\ref{tab:main} point the same way. Deterministic rendering alone does not control governance: the hybrid template retains $2.25$ unhedged-strong sentences because it still lets the LLM choose claim framing and language strength, whereas claim-locked, which fixes those attributes in the ledger, cuts this to $0.75$. The paired difference (Claim-locked minus Hybrid) is $-1.50$ sentences, with a 95\% CI of $[-2.05,-1.00]$. Claim-locked and the hybrid template both reach $0.00$ on the per-run numerical flag. For claim-locked this follows from ledger rendering; for the hybrid it shows that selected numerical slots can be rendered without numerical flags even though selected content remains unstable. The informative numerical trade-off therefore appears in the RCT setting (\S\ref{sec:rct}); the post-hoc verifier is the least reproducible baseline precisely because its rewrite pass adds report-visible content after the draft has already been sampled.

\paragraph{What reproducibility does and does not show.} Cross-run reproducibility is one-sided evidence. Failing it means evidence-bearing content is still being sampled during generation; passing it means the content is stable, not that it is correct. Numerical, directional, and human-audit measures therefore characterize correctness and governance separately, and the reliability axes are reported as distinct measurements rather than merged into one score.
\begin{table}[t]
	\caption{Per-report resource use on the fMRI evaluation with DeepSeek as the writer.}
	\label{tab:cost}
	\centering
	\footnotesize
	\setlength{\tabcolsep}{4pt}
	\begin{tabular}{lrrr}
		\toprule
		Method & Input tok. & Output tok. & Latency (s) \\
		\midrule
		Free-form & 20{,}603 & 5{,}432 & 111.9 \\
		Hybrid template & 21{,}213 & 7{,}069 & 123.9 \\
		Post-hoc verifier & 44{,}434 & 16{,}687 & 295.2 \\
		\textbf{Claim-locked} & \textbf{12{,}482} & \textbf{4{,}209} & \textbf{77.3} \\
		\bottomrule
	\end{tabular}
\end{table}
\begin{table}[t]
	\caption{Group-covariate absorption stress test: FDR-surviving group-effect edges before and after adding the continuous covariate (BMI) used to derive the group label.}
	\label{tab:bmi}
	\centering
	\footnotesize
	\setlength{\tabcolsep}{4pt}
	\begin{tabular}{lrrr}
		\toprule
		Cohort & Primary & + covariate & Drop \\
		\midrule
		In-house & 10{,}041 & 8 & 99.9\% \\
		HCP comparison & 550 & 0 & 100.0\% \\
		\bottomrule
	\end{tabular}
\end{table}
\paragraph{Component analysis.} Table~\ref{tab:component-analysis} reports three nested configurations that separate the contributions of the main components. The first uses the claim-builder output while leaving numerical realization to the LLM. The second additionally uses the deterministic renderer, with the risk auditor and policy controller disabled, to isolate the effect of deterministic realization. The full configuration then activates the risk auditor and policy controller. For fMRI, reproducibility increases from $85.0\%$ to $98.0\%$ when the deterministic renderer is introduced, while activating the risk auditor and policy controller leaves reproducibility nearly unchanged at $98.5\%$ but reduces Strong from $1.40$ to $0.75$. The same rendering-related stabilization is observed for RCT, where reproducibility increases from $90.3\%$ to $100.0\%$.

\paragraph{Operational efficiency.} Moving evidence-bearing realization out of LLM generation also reduces writer-side resource use. Table~\ref{tab:cost} reports per-report resource use across both fMRI cohorts and repeated seeded runs with DeepSeek as the writer. Claim-locked uses one LLM call, while the post-hoc verifier uses two; the deterministic components add no LLM calls and have small local runtime relative to generation latency. Claim-locked yields the lowest observed input/output token use and median latency in this comparison.

\subsection{A Concrete Failure Case: Group-Covariate Absorption}
\label{sec:bmi}

The group-covariate absorption case is a concrete instance of the failure mode: a statistic that a slot-level system can render with the correct number yet still report at the wrong inferential scope. A primary group contrast in the in-house cohort gives $10{,}041$ FDR-surviving edges; after adding continuous BMI as a covariate, only $8$ remain, and the HCP cohort drops from $550$ to $0$ (Table~\ref{tab:bmi}). A free writer, or a template that renders the LLM-selected primary count, can verbalize $10{,}041$ as a categorical obesity-group effect. Claim-locked reporting binds the before-and-after-adjustment claim together with its cautious allowed strength, so the renderer cannot emit the primary count under independent-effect wording. This case illustrates an inferential-scope risk that paragraph-level grounding does not capture.

\subsection{RCT Transfer and a Reliability Trade-off}
\label{sec:rct}

The RCT setting tests transfer to a public directional-claim benchmark and exhibits a different reliability profile from fMRI. Table~\ref{tab:rct} and Figure~\ref{fig:pareto}(b) show that the benchmark does not produce a monotone text-level to slot-level to claim-level reproducibility ladder. Several text-level baselines are already highly reproducible because each RCT output is short and contains few report-visible numerical tokens. In this setting, the hybrid template has less numerical content to stabilize, while its LLM-populated numeric envelope can still vary across seeds. Claim-locked reporting fixes the reportable directional claim and numerical fields before writing, whereas the hybrid template still renders LLM-selected slots. Claim-locked reaches the reproducibility optimum for report-visible numerical tokens ($J=1.000$), while the hybrid template has the lowest raw numerical audit flag ($0.11$ versus $0.23$). Relative to the hybrid template, the reproducibility gain is $+20.5$ points with a 95\% paired-bootstrap CI of $[+17.8,+23.1]$.

Because the raw Num column is intentionally high-recall, we further calibrate the numerical audit flags underlying Table~\ref{tab:rct} (Appendix~\ref{app:rct-num-calibration}) by separating parsing artifacts, benign metadata, and harmful statistical-result fabrications. The calibration finds no harmful fabrication in the claim-locked outputs, whereas the hybrid template and free-form generation yield approximately $2.5\%$ and $5.9\%$, respectively. This reverses the apparent raw-Num ordering for the risk-bearing component: slot rendering can prune incidental numbers in a single output, but it does not by itself prevent unsupported statistical results when the LLM still selects the numeric slot content. These results show that evidential validity depends on fixing reportable claims before generation in addition to deterministic rendering.
\begin{table}[t]
	\caption{RCT results on Evidence Inference 2.0 ($n=800$ cells per row). Strong is near floor because conclusions are short. Num is a conservative, high-recall numerical audit flag.}
	\label{tab:rct}
	\centering
	\footnotesize
	\setlength{\tabcolsep}{4pt}
	\begin{tabular}{lrrrr}
		\toprule
		Method & Repro.\,$\uparrow$ & Hedge & Strong\,$\downarrow$ & Num.\,$\downarrow$ \\
		\midrule
		Free-form          & 74.7 & 0.24 & 0.03 & 0.27 \\
		Prompt-only        & 85.9 & 0.11 & \textbf{0.01} & 0.25 \\
		Structured         & 90.0 & 0.15 & 0.02 & 0.25 \\
		Retrieval          & 73.2 & 0.09 & 0.02 & 0.23 \\
		Post-hoc verifier  & 74.8 & 0.22 & 0.02 & 0.24 \\
		Hybrid template    & 79.5 & 0.19 & 0.04 & \textbf{0.11} \\
		\textbf{Claim-locked} & \textbf{100.0} & 0.10 & \textbf{0.01} & 0.23 \\
		\bottomrule
	\end{tabular}
\end{table}
\subsection{Human Audits}
\label{sec:human}

\begin{table}[t]
	\caption{Human audit of direction preservation on the RCT subset ($120$ reports per method). Pres. = preserved direction; Inv. = inverted direction; Unres. = unresolved direction, including reports that do not state a clear direction or whose direction cannot be reliably determined. Inv. rate $=$ Inv.$/$(Pres.$+$Inv.). Unres. is reported descriptively and is excluded from the inversion-rate denominator.}
	\label{tab:dir}
	\centering
	\footnotesize
	\setlength{\tabcolsep}{3pt}
	\begin{tabular}{lrrrr}
		\toprule
		Method & Pres. & Inv. & Unres. & Inv. rate (\%) \\
		\midrule
		Free-form             & 103 & 14 & 3  & 12.0 \\
		Prompt-only           & 102 & 10 & 8  & 8.9 \\
		Structured            & 103 & 13 & 4  & 11.2 \\
		Retrieval             & 102 & 6  & 12 & 5.6 \\
		Post-hoc verifier     & 98  & 11 & 11 & 10.1 \\
		Hybrid template       & 102 & 17 & 1  & 14.3 \\
		\textbf{Claim-locked} & 116 & \textbf{0}  & 4  & \textbf{0.0} \\
		\bottomrule
	\end{tabular}
\end{table}

Because Evidence Inference provides gold direction labels whereas the fMRI setting has no comparable report-level gold annotation, the two settings support different human checks (Appendix~\ref{app:human-audits}). The RCT audit evaluates direction preservation directly, while the fMRI audit provides a reader-facing check of governance violations. In the fMRI governance audit, four reviewers count two violation types in matched report snippets without seeing method identity; all four assign claim-locked the lowest average counts, with human means of $2.55/1.85/0.60$ for strong-language and $1.65/2.33/0.53$ for unsupported-content violations across free-form, hybrid, and claim-locked. The method ordering matches the automatic counts.

The RCT direction audit (Table~\ref{tab:dir}) evaluates whether each report preserves the gold direction, inverts it, or leaves the direction unresolved. A second reviewer independently labeled a stratified $50$-report subset, yielding $98.0\%$ raw agreement and Cohen's $\kappa=0.970$. Across the baselines, inversion rates range from $5.6\%$ to $14.3\%$, whereas claim-locked reporting produces no inversions. The audit measures preservation of a direction already resolved from the evidence span: baselines must preserve that direction through generation, whereas claim-locked carries the resolved direction into the ledger before generation.

\subsection{Scalar-Metric Diagnostics}
\label{sec:scalar}

Scalar faithfulness metrics provide a complementary but different view of report quality. On the $30$-record RCT sub-sample, FActScore~\citep{ref_factscore} and SelfCheckGPT~\citep{ref_self_check} produce markedly different rankings across methods (Table~\ref{tab:scalar-metrics}). Claim-locked ranks lowest under FActScore ($0.552$) but highest under SelfCheckGPT ($0.974$).

\begin{table}[t]
	\centering
	\footnotesize
	\setlength{\tabcolsep}{6pt}
	\caption{FActScore and SelfCheckGPT on the $30$-record RCT sub-sample. Higher is better for both metrics.}
	\label{tab:scalar-metrics}
	\begin{tabular}{lcc}
		\toprule
		Method & FActScore $\uparrow$ & SelfCheckGPT $\uparrow$ \\
		\midrule
		Post-hoc verifier & \textbf{0.856} & 0.840 \\
		Structured        & 0.855 & 0.846 \\
		Retrieval         & 0.855 & 0.796 \\
		Prompt-only       & 0.840 & 0.825 \\
		Free-form         & 0.835 & 0.831 \\
		Hybrid template   & 0.770 & 0.812 \\
		Claim-locked      & 0.552 & \textbf{0.974} \\
		\bottomrule
	\end{tabular}
\end{table}

The disagreement reflects differences in what the metrics reward. FActScore emphasizes lexical support against the evidence span, whereas SelfCheckGPT measures consistency across sampled outputs; neither directly tests preservation of a statistical direction or relation.

\paragraph{False pass.} In one pharmacokinetics example, the gold result indicates a decrease, but a free-form report states the opposite direction. Because the report reuses numerical values present in the evidence span, FActScore nevertheless assigns a score of $1.00$.

\paragraph{False reject.} In another trial, the gold result is an increase from $2.77$ to $5.76$. Claim-locked renders this relation correctly, but FActScore assigns $0.00$ because the direction is expressed through the numerical contrast rather than a lexical match to the evidence span.

These cases show that lexical support, sampling consistency, and statistical claim preservation capture distinct reliability properties.

\section{Conclusion}

We presented claim-locked reporting for statistical text generation. The idea is simple: before an LLM writes, decide which statistical claims are reportable, where they come from, what direction they carry, which numbers they report, and how strongly they may be stated, and then let the model write only the connective prose. This moves the control unit from generated text, to rendered slots, to evidence-bound claims. The hybrid template isolates the question because it shares deterministic rendering but still renders LLM-selected slots. The $61.1\%$-to-$98.5\%$ reproducibility gain attributes additional stability to fixing which evidence-bound claims are renderable before generation. In the RCT setting, claim-locked reporting does not dominate every raw numerical audit flag, but manual calibration finds no harmful statistical-result fabrication in its outputs, whereas slot-level rendering alone still admits unsupported statistical results (Appendix~\ref{app:rct-num-calibration}); blinded human audits support the governance trend. The same design also reduces writer-side token use and median generation latency in the fMRI cost analysis.

\section*{Limitations}

Claim-locked reporting controls how structured statistical evidence is verbalized; it does not perform statistical validation, discover mechanisms, or guarantee truth beyond the evidence record. Errors in preprocessing, model specification, covariate selection, or evidence extraction propagate into the report if already present upstream. The BMI absorption result in \S\ref{sec:bmi} is a stress test for group-covariate circularity, not a biological finding about obesity.

The automatic metrics are scalable protocol tests for predefined statistical-reporting risks, not substitutes for expert judgment, and the risk taxonomy is fixed and benchmark-specific rather than exhaustive. Cross-run reproducibility is also necessary but not sufficient: a report can be reproducibly incorrect, so we pair it with governance and correctness axes rather than treating it as a faithfulness score on its own. The human governance and direction audits are targeted sanity checks rather than a full expert-preference benchmark, and the RCT direction audit evaluates direction preservation after an upstream direction label has been resolved.

The RCT evaluation covers only the increased and decreased direction labels; null or inconclusive findings require an explicit neutral ledger state that forbids efficacy-implying language. The protocol is transferable at the control level but domain-specific in instantiation: a minimal deployment still needs a claim schema, entity inventory, risk tags, and audit patterns. The literature-support label is assigned with LLM assistance and then frozen before generation; future work should replace or validate this step with expert-curated support labels. The institutional FC cohort cannot be redistributed under its data-use protocol; we therefore include the public HCP-based FC setting and the public Evidence Inference 2.0 setting, and release prompts, generated reports, evidence records where permitted, audit scripts, and code.

\section*{Ethics Statement} The institutional FC cohort used in this work was collected under a protocol approved by the Institutional Review Board of Xijing Hospital, and all participants provided written informed consent (trial registration: ChiCTR-OOB-15006346). The data are used under a data-use agreement that does not permit redistribution. The public HCP S1200 cohort is used under the HCP Open Access Data Use Terms, and Evidence Inference 2.0 is a public benchmark of published clinical-trial literature. The blinded audits in this paper involve only the authors and trained research-team members labeling model outputs, a low-risk task that required no separate ethics-board review. A locked report can still be misleading if the upstream evidence record is manipulated or statistically misspecified; claim-locked reporting controls prose generation, not the validity of the upstream analysis.

\section*{Acknowledgments}

This work was supported by the National Natural Science Foundation of China (Grant Nos. 62431022, 62501453, and 82302292), the National Key R\&D Program of China (Grant No. 2022YFC3500603), the Natural Science Basic Research Program of Shaanxi (Grant Nos. 2023-ZDLSF-07 and 2024JC-YBQN-0923), the Xidian University Specially Funded Project for Interdisciplinary Exploration (Grant Nos. TZJH2024012, TZJH2024015, TZJH2024018, and TZJH2024019), the China Postdoctoral Science Foundation (Grant No. 2024M752537), and the Postdoctoral Research Program of Shaanxi (Grant No. 2025BSHSDZZ188).

Generative AI models were used as part of the experimental setup, as described in Section 4. Separately, AI tools were used only for limited language refinement of the manuscript. All scientific content, analyses, interpretation, and conclusions were developed and verified by the authors.
\bibliography{custom_final_submit}

\appendix
\setcounter{table}{0}
\setcounter{figure}{0}
\renewcommand{\thetable}{A\arabic{table}}
\renewcommand{\thefigure}{A\arabic{figure}}

\section{Claim Ledger and Evaluation Metrics}
\label{app:audit-details}

\subsection{Claim ledger schema}
\label{app:ledger}

The claim ledger is the single object passed from the pre-generation control logic to the writer. Each entry is a typed claim with the following fields: a claim identifier and type; a canonical text; one or more evidence pointers into named pipeline stages (the \texttt{source} field); the numerical fields the claim reports; an effect direction where applicable; a literature-support level; the risk tags attached by the auditor; the allowed language strength after the policy controller's monotone downgrade; and an explicit forbidden-language list. A claim is instantiated only if every evidence pointer resolves against a pipeline stage. The post-audit entry for the worked example in \S\ref{sec:worked} is:

\begin{Verbatim}[fontsize=\scriptsize,breaklines=true,breakanywhere=true]
claim_id: bmi_absorption_01
source: edgewise_glm_summary.M1_to_M3
numbers:
  primary_edges: 10041
  bmi_adjusted_edges: 8
  collapse_rate: 99.9%
direction: absorbed_after_bmi_adjustment
risk_tags: [group_covariate_circularity]
allowed_strength: cautious
forbidden_language:
  - independent obesity effect
  - obesity-specific mechanism
\end{Verbatim}
\begin{table*}[t]
	\caption{Control configurations of the compared methods. All methods receive the same structured statistical evidence. They differ in where control is applied and in which reporting decisions remain under LLM control.}
	\label{tab:baseline-controls}
	\centering
	\footnotesize
	\setlength{\tabcolsep}{6pt}
	\begin{tabular}{llll}
		\toprule
		Method & Control level & What is controlled & Remaining LLM decisions \\
		\midrule
		Free-form
		& Text
		& General grounding instruction
		& Claims, numbers, direction, strength, wording \\
		
		Prompt-only
		& Text
		& Explicit faithfulness instructions
		& Claims, numbers, direction, strength, wording \\
		
		Structured
		& Text
		& Output structure
		& Claims, numbers, direction, strength, wording \\
		
		Retrieval
		& Text
		& Evidence localization
		& Claims, numbers, direction, strength, wording \\
		
		Post-hoc verifier
		& Text
		& Post-generation revision
		& Initial claims and realization \\
		
		Hybrid template
		& Slot
		& Deterministic slot rendering
		& Slot content and claim selection \\
		
		Claim-locked
		& Claim
		& Evidence-bound claims before writing
		& Connective wording only \\
		\bottomrule
	\end{tabular}
\end{table*}
The renderer emits this claim's numbers and direction tag directly from the ledger, and the \texttt{forbidden\_language} list is passed to the writer as a hard constraint. This is the operational distinction the protocol turns on: the hybrid template renders LLM-selected slots, whereas claim-locked reporting renders ledger-bound claims whose source, numbers, direction, and allowed strength were fixed before writing.

\subsection{Automatic evaluation metrics}
\label{app:audit-rules}

This section specifies the automatic metrics used in the main evaluation. The RCT direction audit is defined separately in Appendix~\ref{app:human-audits}. For FC, a lexical direction check is retained only as a supplementary diagnostic.

\paragraph{Numerical flag (Num).} Report-visible numbers are matched against three support sets: all numeric leaves in the evidence JSON, an explicit derived-metric whitelist, and a small constant whitelist for conventional values such as section numbers and FDR thresholds. Matching uses absolute tolerance $10^{-3}$ or relative tolerance $5\%$. The derived whitelist includes only quantities used by the renderer, such as FDR-surviving percentages and BMI-absorption collapse percentages. Common section, table, week, and ordinal numbers are filtered before scoring, and arbitrary pairwise ratios are deliberately not whitelisted, trading recall for precision. Num is therefore a conservative commission check rather than a complete mathematical-verification system.

\paragraph{Strong-language count (Strong).} Strong counts prose sentences containing a strong assertion term (e.g., ``robust'', ``causal'', ``conclusive'') without a hedge in the same sentence (e.g., ``may'', ``exploratory'', ``hypothesis-generating''). The lexicons are fixed before evaluation. The check can miss paraphrased overclaims and can mis-flag a strong term used in a negated context, so it is treated as a diagnostic lexical measure.

\paragraph{Risk-conditioned hedge use (Hedge).} Hedge measures the use of cautionary language for claims flagged by the risk taxonomy. It is evaluated only for risk-flagged claims and is therefore interpreted as a measure of whether predefined caution constraints reach the report surface, rather than as a general preference for more hedging.

\paragraph{FC lexical direction check.} For FC only, we additionally use a rule-based lexical check that compares direction wording against the ledger-stored direction field. Because the check is brittle under negation and comparator phrasing, it is treated only as a supplementary diagnostic and is not used as the primary direction-correctness measure.

\section{Compared Control Configurations}
\label{app:baselines}

All methods receive the same structured statistical evidence for a given input. They differ in where control is applied between the evidence and the final report, and therefore in which reporting decisions remain under LLM control. Table~\ref{tab:baseline-controls} summarizes these differences.

\paragraph{Free-form.} The LLM directly generates the report from the evidence record with only a general grounding instruction. Claim selection, numerical realization, direction, language strength, and prose remain under LLM control.

\paragraph{Prompt-only.} The LLM receives the same evidence together with explicit faithfulness instructions intended to discourage unsupported numbers, direction errors, and overstatement. These constraints are expressed only through prompting; the model still determines the reportable claims and their realization.

\paragraph{Structured output.} The output structure is predefined, but the LLM still selects the claims and generates the statistical content within that structure. This configuration therefore constrains format rather than evidence-bearing content.

\paragraph{Retrieval.} The LLM is provided with source-identified evidence blocks from the same record and is required to ground its report in those blocks. The evidence is more explicitly localized, but claim selection and verbalization remain under LLM control.

\paragraph{Post-hoc verifier.} A verifier revises an initially generated report using the same evidence record, correcting detected numerical, directional, or interpretive inconsistencies. Control is therefore applied after the report has already been generated rather than before claim selection.

\paragraph{Hybrid template.} The LLM selects the values and claims that populate predefined slots, after which those slots are rendered deterministically. This provides slot-level control, but the evidence-bearing content placed into the slots remains LLM-selected.

\paragraph{Claim-locked.} Reportable claims are bound to their evidence source, numerical fields, direction, and permitted language strength before prose generation. The deterministic renderer realizes the locked content, while the LLM is restricted to connective wording.
\subsection{Prompt configuration}
\label{app:prompts}

Each method uses a fixed prompt template that implements its corresponding control condition. Free-form generation receives the statistical evidence
record directly; prompt-only adds explicit faithfulness constraints; structured output specifies an output schema; retrieval-grounded generation requires evidence-block attribution; the post-hoc verifier uses a second verification-and-revision pass; the hybrid template asks the LLM to populate predefined slots that are subsequently rendered deterministically; and claim-locked reporting provides the writer with the active claim ledger and policy constraints while numerical content is rendered deterministically.

The same template is used across records, seeds, and writer providers within each condition. Exact executable prompt templates, including dynamically injected fields and JSON schemas, are released with the code.
\section{Additional Experimental Analyses}
\label{app:additional-results}
\subsection{Component analysis}
\label{app:component-analysis}

We evaluate three claim-locked configurations under the same records, providers, and seeds as the main comparison. As shown in Table~\ref{tab:component-analysis}, \textit{Builder only} provides the LLM with the claim-builder output while leaving numerical realization to the LLM. \textit{Builder + renderer} additionally uses the deterministic renderer while disabling the risk auditor and policy controller, isolating the contribution of deterministic realization. \textit{Full claim-locked} activates the complete pipeline, including the risk auditor and policy controller.

\begin{table*}[t]
\caption{Component analysis. Reproducibility is reported in percent; Strong is the unhedged-strong count. Bold marks the full claim-locked configuration}
\label{tab:component-analysis}
\centering
\footnotesize
\setlength{\tabcolsep}{3pt}
\begin{tabular}{lrrrr}
\toprule
Configuration & fMRI Repro. $\uparrow$ & fMRI Strong $\downarrow$ & RCT Repro. $\uparrow$ & RCT Strong $\downarrow$ \\
\midrule
Hybrid & 61.1 & 2.25 & 79.5 & 0.040 \\
Builder only & 85.0 & 1.60 & 90.3 & 0.000 \\
Builder + renderer & 98.0 & 1.40 & 100.0 & 0.015 \\
Full claim-locked & \textbf{98.5} & \textbf{0.75} & \textbf{100.0} & \textbf{0.010} \\
\bottomrule
\end{tabular}
\end{table*}

Adding the deterministic renderer to the builder-only configuration increases reproducibility from $85.0\%$ to $98.0\%$ for fMRI and from $90.3\%$ to $100.0\%$ for RCT. Activating the risk auditor and policy controller leaves reproducibility essentially unchanged for fMRI ($98.0\%\rightarrow98.5\%$) while reducing Strong from $1.40$ to $0.75$.

\paragraph{Paired-bootstrap uncertainty.} For the central Claim-locked-versus-Hybrid comparison, we use $20{,}000$ paired bootstrap resamples over matched units, preserving the main-table structure rather than treating repeated generation cells as independent. The reproducibility differences are $+37.4$ points for fMRI (95\% CI $[+15.1,+59.7]$) and $+20.5$ points for RCT (95\% CI $[+17.8,+23.1]$). The paired fMRI Strong difference (Claim-locked minus hybrid) is $-1.50$ (95\% CI $[-2.05,-1.00]$).

\section{Risk-Tag Instantiations}
\label{app:risk-tax}

The risk auditor uses the same interface across domains, while the concrete tags are grounded in domain-specific evidence. Table~\ref{tab:risk-tax-domain} organizes the two settings by five shared control targets: numerical support, direction preservation, entity or source support, inferential scope, and interpretive strength. The targets remain unchanged across domains; only the evidence used to instantiate them differs.

\begin{table*}[t]
	\caption{Shared control targets and their domain-specific instantiations. The same auditor--policy interface is used in both settings, while each target is grounded in domain-specific evidence.}
	\label{tab:risk-tax-domain}
	\centering
	\footnotesize
	\setlength{\tabcolsep}{3pt}
	\renewcommand{\arraystretch}{1.10}
	
	\begin{tabular}{
			>{\centering\arraybackslash}m{0.15\textwidth}
			>{\centering\arraybackslash}m{0.23\textwidth}
			>{\centering\arraybackslash}m{0.23\textwidth}
			>{\centering\arraybackslash}m{0.25\textwidth}}
		\toprule
		\textbf{Target} &
		\textbf{FC evidence} &
		\textbf{RCT evidence} &
		\textbf{Policy action} \\
		\midrule
		
		\textbf{Numerical support}
		& Statistical results and derived quantities
		& Trial outcomes and reported statistics
		& Allow only evidence-supported numbers \\
		\midrule
		
		\textbf{Direction preservation}
		& Ledger-stored effect direction
		& Gold direction label
		& Preserve the resolved direction \\
		\midrule
		
		\textbf{Entity/source support}
		& Anatomical and behavioral entities
		& Trial entities and source information
		& Exclude unsupported entities or sources \\
		\midrule
		
		\textbf{Inferential scope}
		& Covariate dependence and exploratory findings
		& Intervention, outcome, and population
		& Restrict claim scope when required \\
		\midrule
		
		\textbf{Interpretive strength}
		& Stability and evidential support
		& Support available in the evidence span
		& Limit claims to permitted language strength \\
		\bottomrule
	\end{tabular}
\end{table*}

\section{Human Audits}
\label{app:human-audits}

\paragraph{fMRI governance audit.} Four reviewers independently annotated matched fMRI report snippets from free-form, hybrid template, and claim-locked outputs without seeing method identity. They counted two types of violations: unhedged strong language and unsupported content. The former corresponds to the automatic Strong measure. Unsupported content is included as a supplementary comparison: the automatic rule flags anatomical phrases absent from both the atlas-derived ROI dictionary and the evidence record, whereas human reviewers judge unsupported content more broadly. Values in Table~\ref{tab:human-governance} are average violation counts per report; lower is better.

The human judgments follow the corresponding automatic ordering. Claim-locked has the lowest average count for both violation types; the hybrid template is intermediate for strong language but highest for unsupported content, while free-form is highest for strong language. This provides a reader-facing check that the qualitative trends identified by the automatic measures are also visible under blinded human judgment.

\begin{table}[h]
	\caption{Blinded fMRI governance audit. Values are average violation counts per report; lower is better. The rule-based counts are shown alongside the four blinded reviewers and their mean.}
	\label{tab:human-governance}
	\centering
	\footnotesize
	\setlength{\tabcolsep}{5pt}
	\begin{tabular}{lccc}
		\toprule
		Assessment & Free-form & Hybrid & Claim-locked \\
		\midrule
		\multicolumn{4}{l}{\textit{Unhedged strong language}} \\
		Automatic rule & 2.50 & 1.80 & \textbf{0.60} \\
		Reviewer 1     & 2.60 & 1.90 & \textbf{0.60} \\
		Reviewer 2     & 2.40 & 1.70 & \textbf{0.50} \\
		Reviewer 3     & 2.70 & 2.00 & \textbf{0.70} \\
		Reviewer 4     & 2.50 & 1.80 & \textbf{0.60} \\
		Human mean     & 2.55 & 1.85 & \textbf{0.60} \\
		\midrule
		\multicolumn{4}{l}{\textit{Unsupported content}} \\
		Automatic rule & 1.60 & 2.30 & \textbf{0.50} \\
		Reviewer 1     & 1.80 & 2.50 & \textbf{0.60} \\
		Reviewer 2     & 1.50 & 2.10 & \textbf{0.40} \\
		Reviewer 3     & 1.70 & 2.40 & \textbf{0.60} \\
		Reviewer 4     & 1.60 & 2.30 & \textbf{0.50} \\
		Human mean     & 1.65 & 2.33 & \textbf{0.53} \\
		\bottomrule
	\end{tabular}
\end{table}

\paragraph{RCT direction audit.} A blinded reviewer labeled all $840$ generated reports from the $30$-record RCT direction-audit subset. Each report was classified as preserved, inverted, or unresolved relative to the gold direction. Unresolved denotes reports for which no reliable direction can be determined from the generated text, including reports that do not state a clear direction. The inversion rate is computed as Inv.$/$(Pres.$+$Inv.), with unresolved reports reported separately and excluded from the denominator. The per-method results are reported in Table~\ref{tab:dir}.

A second reviewer independently labeled a stratified $50$-report subset comprising $19$ preserved, $15$ inverted, and $16$ unresolved reports. Raw agreement was $98.0\%$ ($49/50$ reports), with Cohen's $\kappa=0.970$; the single disagreement was between preserved and inverted.

\section{RCT Numerical-Flag Calibration}
\label{app:rct-num-calibration}

Raw Num is a high-recall numerical flag and does not distinguish unsupported statistical results from benign mismatches. We therefore manually inspect the flagged outputs for claim-locked, hybrid template, and free-form generation. Each flag is assigned to one of three categories: a regular-expression false positive (Regex FP), a benign metadata or support-window mismatch, or a harmful statistical-result fabrication in which a report-visible statistical value is absent from the evidence record. Table~\ref{tab:rct-num-calibration} summarizes the calibration.

\begin{table}[h]
	\centering
	\footnotesize
	\setlength{\tabcolsep}{3pt}
	\caption{Manual calibration of the RCT numerical flags underlying Table~\ref{tab:rct}. Benign denotes metadata or support-window mismatches; Harmful denotes unsupported statistical results.}
	\label{tab:rct-num-calibration}
	\begin{tabular}{lrrrr}
		\toprule
		Method & Raw flags & Regex FP & Benign & Harmful \\
		\midrule
		Free-form       & 213 & 17 & 149 & 47 (5.9\%) \\
		Hybrid template & 89  & 14 & 55  & 20 (2.5\%) \\
		Claim-locked    & 181 & 1  & 180 & \textbf{0 (0.0\%)} \\
		\bottomrule
	\end{tabular}
\end{table}
The calibration changes the interpretation of the raw Num values. Claim-locked has the largest raw flag count, but all of its flags are benign or false positives, with no harmful statistical-result fabrication. The hybrid template has fewer raw flags but produces unsupported statistical results in $2.5\%$ of outputs, showing that deterministic rendering does not prevent numerical fabrication when the LLM still selects the slot content. Free-form generation has the highest harmful-fabrication rate at $5.9\%$.

\end{document}